\documentclass{wscpaperproc}
\usepackage{latexsym}
\usepackage{graphicx}
\usepackage{mathptmx}
\usepackage[T1]{fontenc}
\usepackage[draft]{changes}

\usepackage[normalem]{ulem}

\usepackage{amsmath}
\usepackage{amsfonts}
\usepackage{amssymb}
\usepackage{amsbsy}
\usepackage{amsthm}
\usepackage{subfigure}
\usepackage{multirow}
\usepackage{makecell}

\usepackage[pdftex,colorlinks=true,urlcolor=blue,citecolor=black,anchorcolor=black,linkcolor=black]{hyperref}

\newtheoremstyle{wsc}
{3pt}
{3pt}
{}
{}
{\bf}
{}
{.5em}
{}

\theoremstyle{wsc}

\begin{document}

%
%

\pagestyle{fancyplain}

\thispagestyle{plain}
\firstPageHead{}

\chead{\fancyplain{}{\itshape Cheng, Luo, Fan, Zhang, and Li}}

\rhead{}
\cfoot{}
\renewcommand{\headrulewidth}{0pt} 

\makeatletter
\let\@internalcite\cite
\def\cite{\def\@citeseppen{-1000}%
    \def\@cite##1##2{(##1\if@tempswa , ##2\fi)}%
    \def\citeauthoryear##1##2##3{##1 ##3}\@internalcite}
\def\citeNP{\def\@citeseppen{-1000}%
    \def\@cite##1##2{##1\if@tempswa , ##2\fi}%
    \def\citeauthoryear##1##2##3{##1 ##3}\@internalcite}
\def\citeN{\def\@citeseppen{-1000}%
    \def\@cite##1##2{##1\if@tempswa, ##2)\else{}\fi}%
    \def\citeauthoryear##1##2##3{##1 (##3)}\@citedata}
\def\citeA{\def\@citeseppen{-1000}%
    \def\@cite##1##2{(##1\if@tempswa , ##2\fi)}%
    \def\citeauthoryear##1##2##3{##1}\@internalcite}
\def\citeANP{\def\@citeseppen{-1000}%
    \def\@cite##1##2{##1\if@tempswa , ##2\fi}%
    \def\citeauthoryear##1##2##3{##1}\@internalcite}
\def\shortcite{\def\@citeseppen{-1000}%
    \def\@cite##1##2{(##1\if@tempswa , ##2\fi)}%
    \def\citeauthoryear##1##2##3{##2 ##3}\@internalcite}
\def\shortciteNP{\def\@citeseppen{-1000}%
    \def\@cite##1##2{##1\if@tempswa , ##2\fi}%
    \def\citeauthoryear##1##2##3{##2 ##3}\@internalcite}
\def\shortciteN{\def\@citeseppen{-1000}%
    \def\@cite##1##2{##1\if@tempswa, ##2\else{}\fi}%
    \def\citeauthoryear##1##2##3{##2 (##3)}\@citedata}
\def\shortciteA{\def\@citeseppen{-1000}%
    \def\@cite##1##2{(##1\if@tempswa , ##2\fi)}%
    \def\citeauthoryear##1##2##3{##2}\@internalcite}
\def\shortciteANP{\def\@citeseppen{-1000}%
    \def\@cite##1##2{##1\if@tempswa , ##2\fi}%
    \def\citeauthoryear##1##2##3{##2}\@internalcite}
\def\citeyear{\def\@citeseppen{-1000}%
    \def\@cite##1##2{(##1\if@tempswa , ##2\fi)}%
    \def\citeauthoryear##1##2##3{##3}\@citedata}
\def\citeyearNP{\def\@citeseppen{-1000}%
    \def\@cite##1##2{##1\if@tempswa , ##2\fi}%
    \def\citeauthoryear##1##2##3{##3}\@citedata}
%
%
%
\def\@citedata{%
    \@ifnextchar [{\@tempswatrue\@citedatax}%
                  {\@tempswafalse\@citedatax[]}%
}

\def\@citedatax[#1]#2{%
\if@filesw\immediate\write\@auxout{\string\citation{#2}}\fi%
  \def\@citea{}\@cite{\@for\@citeb:=#2\do%
    {\@citea\def\@citea{, }\@ifundefined
       {b@\@citeb}{{\bf ?}%
       \@warning{Citation `\@citeb' on page \thepage \space undefined}}%
{\csname b@\@citeb\endcsname}}}{#1}}%

%
\def\@citex[#1]#2{%
\if@filesw\immediate\write\@auxout{\string\citation{#2}}\fi%
  \def\@citea{}\@cite{\@for\@citeb:=#2\do%
    {\@citea\def\@citea{; }\@ifundefined
       {b@\@citeb}{{\bf ?}%
       \@warning{Citation `\@citeb' on page \thepage \space undefined}}%
{\csname b@\@citeb\endcsname}}}{#1}}%

%
\def\@biblabel#1{}
\makeatother



\newdimen\bibindent
\bibindent=0.0em
\def\thebibliography#1{\section*{\refname}\list
   {}{\settowidth\labelwidth{[#1]}
   \leftmargin\parindent
   \itemindent -\parindent
   \listparindent \itemindent
   \itemsep 0pt
   \parsep 0pt}
   \def\newblock{}
   \sloppy
   \sfcode`\.=1000\relax}


\setlength{\baselineskip}{12.7pt}

\author{
Liqiang Cheng\\
Jun Luo\\[12pt]
	Shanghai Jiao Tong University\\
	NO.1954 Huashan Rd\\
        Shanghai, 200030, CHINA\\
\and
Weiwei Fan\\
\\[12pt]
Tongji University\\
No.1239 Siping Rd\\
Shanghai, 200092, CHINA\\
\and
Yidong Zhang\\
Yuan Li\\[12pt]
Dchain Department, Alibaba Group\\
No.969 West Wen Yi Rd\\
Hangzhou, 311121, CHINA
}

\title{A Deep Q-Network Based on Radial Basis Functions \\
for Multi-echelon Inventory Management}
\maketitle

\section*{ABSTRACT}

This paper addresses a multi-echelon inventory management problem with a complex network topology where deriving optimal ordering decisions is difficult. Deep reinforcement learning (DRL) has recently shown potential in solving such problems, while designing the neural networks in DRL remains a challenge. In order to address this, a DRL model is developed whose Q-network is based on radial basis functions. The approach can be more easily constructed compared to classic DRL models based on neural networks, thus alleviating the computational burden of hyperparameter tuning. Through a series of simulation experiments, the superior performance of this approach is demonstrated compared to the simple base-stock policy, producing a better policy in the multi-echelon system and competitive performance in the serial system where the base-stock policy is optimal. In addition, the approach outperforms current DRL approaches.

\section{INTRODUCTION}
\label{sec:intro}

Supply chain management plays a crucial role in business operations, with inventory management being a core process within it. The focus of this paper is on the multi-echelon inventory ystem which consists of multiple stages or echelons that hold inventory \shortcite{2022Can}, because of its rising popularity in real-world supply chains. For instance, in Alibaba's supply chain, suppliers deliver inventory to a central warehouse, which then allocates inventory to downstream retailers in its region.  In managing such inventory system, the manager may desire to dynamically determine the ordering decision at each period so that  the total supply chain costs is minimized.  The dynamic inventory management problem can be essentially formulated as a Markov Decision Process (MDP).   To address this  problem, many methods have been developed, dating back to \citeN{1960Optimal} and \citeN{1968Metric}. However, these methods are specifically designed for multi-echelon systems with simple structure, e.g.,   the serial system. As pointed by \citeN{2000Foundation}, the optimal inventory policy for the general multi-echelon system is yet unknown, due to the complexity of systems. As a remedy, various approximation methods are proposed, while they often require certain assumptions, such as Poisson process demand or zero lead time. \shortciteN{2022Integrated} investigated an integrated optimization problem of inventory management and transportation vehicle selection. They formulate the problem as a mixed-integer quadratically constrained program and established a convex approximation of the proposed formulation using Cauthy inequalities. More interesting review of these methods can be found  in \citeN{2012Performance}.

Recently, reinforcement learning (RL), also known as approximate dynamic programming (ADP), enjoys notable success in solving MDP.  In a typical RL, an agent consistently interacts with the environment, where at each period, the agent observes the system's state, takes an action and receives  the corresponding reward. Specifically, the action at each period is generated by optimizing the expected value of the total reward starting from the current state, termed by the Q-function.  Classic RL approaches estimate the values of Q-function for all possible action-state pairs and store them in a lookup table named Q-table. Obviously, these methods are not suitable for our inventory management problem, because both our state (i.e., inventory level) and action (i.e., ordering decision) spaces can be large or even continuous.  Instead,  other approaches construct function approximations for the Q-function \shortcite{2011Approximate}.  When the neural networks are used as the function approximator, the corresponding RL is called deep reinforcement learning (DRL) and the constructed approximation is called Q-network. DRL have gained significant attention for achieving human-like intelligence and even surpassing humans in some games, such as Go \shortcite{2021Acquisition} and Atari games \shortcite{2013Playing}.

The appeal of the DRL approach arises from the strong approximation ability of neural networks. In light of this, our paper seeks to adopt the DRL approaches in solving our complicated inventory management problem.

DRL have been applied to solve the inventory management MDPs in different systems. \shortciteN{2021ADeep} proposed a deep Q-network to play the beer game, which is a special serial system. The deep Q-network can achieve near-optimal solutions when playing with teammates who follow a base-stock policy. \shortciteN{2022Solving} developed a double deep Q-network to the lost sales problem, which is a flexible solution that can be applied with different cost parameter settings. \shortciteN{1997Aneuro} derived a neural network dynamic programming approach to solve a two-echelon system, where they manually developed 23 product features to construct the neural network. \shortciteN{2022Can} exploited asynchronous advantage actor-critic algorithm (A3C) for solving lost sales, dual sourcing, and multi-echelon problems. The proposed A3C algorithm can match performance of state-of-the-art heuristics. \shortciteN{2022Multi} applied a multi-agent DRL approach to multi-echelon inventory management, demonstrating that this approach can achieve lower costs and less significant bullwhip effects compared to single-agent DRL methods. They designed a recurrent neural network (RNN) to utilize historical information. Despite the success of DRL in inventory management, designing neural networks in DRL is complex, and tuning hyperparameters remains computationally burdensome \shortcite{2022Can}.

To alleviate the hyperparameter-tuning burden,  we  deploy a deep Q-network approach based on a radial basis functions (RBF) \cite{1988Radial}. The proposed RBF based deep Q-network is a special three-layered neural network, with its hidden layer neurons representing the states of the MDP and the activation function is kernel function. While the hidden layer neurons have a real meaning, the RBF based deep Q-network is easy to design and implement. Our simulation study demonstrates that the proposed RBF based deep Q-network approach achieves appealing performance compared to the base-stock policy and current DRL approaches. For the serial system with one warehouse and one retailer, the RBF based deep Q-network obtains a near-optimal solution (the optimal policy is the base-stock policy). For the multi-echelon system with one warehouse and multiple retailers, the RBF based deep Q-network outperforms the base-stock policy and current DRL approaches.

The structure of this paper is as follows. Section 2 provides the system dynamics and MDP formulation of the multi-echelon inventory management problem considered in this paper. Section 3 introduces our RBF based deep Q-network approach to solve the inventory problem, and Section 4 presents the numerical results obtained from simulated scenarios. Finally, Section 5 concludes this paper.

\section{Multi-echelon Inventory Management}
This section introduces the multi-echelon inventory management model and its corresponding MDP. Section 2.1 presents the dynamics of multi-echelon inventory management, including events that occur and their sequence. Based on the these events, a discrete event simulation model is established. Section 2.2 formulates the MDP for multi-echelon inventory management, and briefly introduces a Q-learning method for solving this MDP, based on which the approach is designed.

\subsection{System Dynamics and Simulation Model}

This section describes the multi-echelon system, which is a one-warehouse multiple-retailer system with $K$ identical retailers. At each period of the infinite periods, random demands materialize at each retailer, and are fulfilled by inventory held at the retailers. Demands are independently and identically distributed through time and among different retailers. The retailers are replenished by a warehouse, while the warehouse is replenished by a supplier. There are delays in the transportation of orders both from the supplier to the warehouse and from the warehouse to each retailer. The delays are considered to be several periods. Hence, the system evolves in discrete time.

\par
\begin{figure}[htb]
{
\centering
\includegraphics[width=0.9\textwidth]{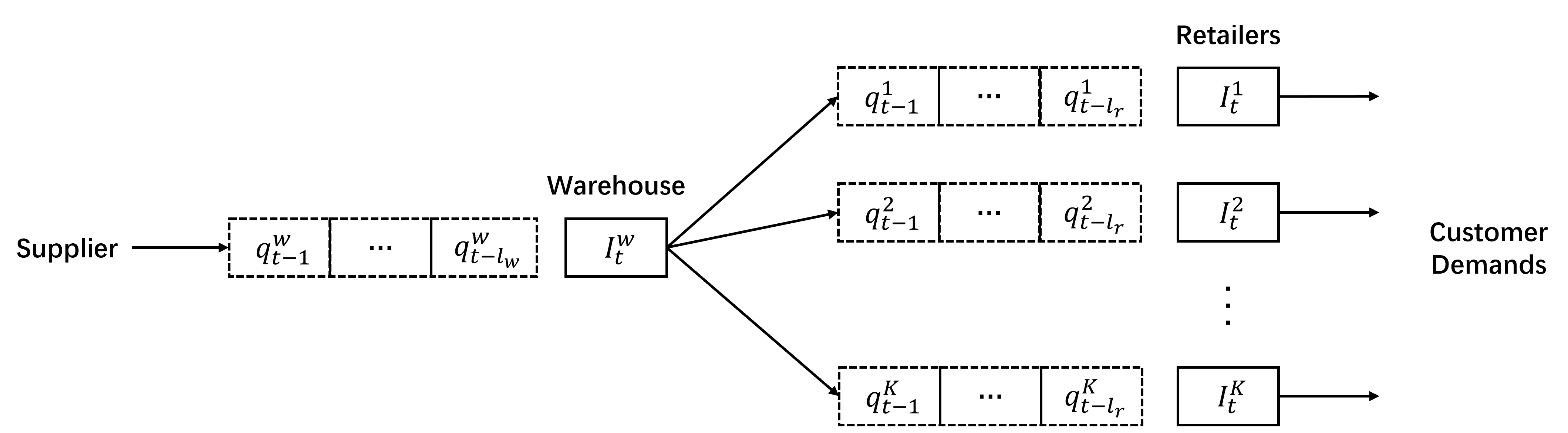}
\caption{Inventory of the multi-echelon system at time point $t$.\label{system}}
}
\end{figure}

 The inventory management is considered over an infinite number of periods. We use time points $t, t+1, \ldots$ to represent the beginning of each period, which also marks the end of the previous period. Without loss of generality, we focus on the multi-echelon inventory management process of one period between time point $t$ and time point $t+1$. At time point $t$, the on-hand inventory of the warehouse and retailers is denoted by $I^w_t$ and $I^i_t$, respectively (where $i=1,\ldots,K$), and the pipeline inventory of the warehouse and retailers is denoted by $Q^w_t=(q^w_{t-1},\ldots,q^w_{t-{l_w}})$ and $Q^i_t=(q^i_{t-1},\ldots,q^i_{t-{l_r}})$ (where $i=1,\ldots,K$). Figure \ref{system} illustrates the on-hand inventory and pipeline inventory at time point $t$, the beginning of one period.

\par The multi-echelon inventory management process between time point $t$ and $t+1$ consists of five events: order arrival, demand fulfillment, special delivery, replenishment, and delivery and update. To provide a more precise description, we introduce virtual time points $t_1$, $t_2$, $t_3$ and $t_4$, at which the first four events occur. The delivery and update event occurs at time point $t+1$, which marks the end of the period. Each of the five events is described in detail below.

\par \textbf{Orders Arrival:} At time point $t_1$, the warehouse and $K$ retailers receive delayed orders that were placed $l_w$ and $l_r$ periods ago, denoted as $q^w_{t-l_w}$ and $q^i_{t-l_r},i=1,\ldots,K$, respectively. These orders are then added to the on-hand inventory of the warehouse and retailers, denoted as $I^w_{t}$ and $I^i_{t}$:
\begin{align*}
I^w_{t_1}&=I^w_{t}+q^w_{t-l_w},\\
I^i_{t_1}&=I^i_{t}+q^i_{t-l_r},1,\ldots,K.
\end{align*}

\par \textbf{Demand Fulfillment:}  At time point $t_2$, each retailer $i$ (where $i=1,\ldots,K$) samples its demand $d^i_t$ from a normal distribution with mean $\mu$ and standard deviation $\sigma$. The retailer then fulfills its demand using its on-hand inventory $I^i_{t_1}$. The on-hand inventory of the warehouse remains unchanged from its value at time point $t_1$ as $I^w_{t_2}=I^w_{t_1}$. While the on-hand inventory of the retailers is updated according to the inventory used to fulfill the demand:
\begin{align*}
I^i_{t_2}&=(I^i_{t_1}-d^i_t)^{+},1,\ldots,K.
\end{align*}

\par \textbf{Special Delivery:} At time point $t_3$, the special delivery event occurs only when the on-hand inventory of a retailer $i$ is insufficient to meet the demand, i.e., $d^i_t>I^i_{t_1}$. In such a case, each unfulfilled demand either waits for a special delivery from the warehouse with probability $P_w$ (if the warehouse on-hand inventory $I^w_{t_2}$ is not $0$) or is lost with probability $1-P_w$, resulting in lost sales. The total number of special deliveries at retailer $i$, denoted as $B^i_t$, follows a binomial distribution $B\left((d^i_t-I^i_{t_1})^+,P_w\right)$. The total number of special deliveries for all retailers is $B_t=\sum^K_{i=1} B^i_t$. During the special delivery event, the on-hand inventory of the retailers remains unchanged, i.e., $I^i_{t_3}=I^i_{t_2}$ for $i=1,\ldots,K$. while the on-hand inventory of the warehouse is updated as follows:
\begin{align*}
I^w_{t_3}&=I^w_{t_2}-B_t.
\end{align*}

\par Any demands that are not fulfilled by the retailers' or warehouse’s on-hand inventory leads to
a shortage cost with a cost rate of $p$. Additionally, special deliveries from the warehouse incur an ordering cost with a cost rate of $c_w$. Therefore, the shortage cost and ordering cost at the period can be calculated as follows:
\begin{align}
    \label{shortage_cost}&\text{shortage cost: }p[\sum_{i=1}^K (d^i_t-I^i_{t_1})^+-B_t],\\
    \label{ordering_cost}&\text{ordering cost: }c_w B_t.
\end{align}

\par \textbf{Replenishment:} At time point $t_4$, the warehouse places an order $q^w_t$, and each retailer $i,i=1,\ldots,K$ places an order $q^i_t$. The total order quantities of the retailers cannot exceed the on-hand inventory of the warehouse, i.e., $\sum_{i=1}^K q^i_t \leq I^w_{t_3}$. Note that both the warehouse and retailers have limited capacities: (1) the maximum order quantity of the warehouse is $C^m$, (2) the warehouse inventory position ${Z}^w_{t_4}=I^w_{t_3}+\sum_{j=1}^{l_w}q^w_{t-j+1}$ cannot exceed $C^w$, and (3) each retailer inventory position ${Z}^i_{t_4}=I^i_{t_3}+\sum_{j=1}^{l_r}q^i_{t-j+1}, i = 1,\ldots,K$, cannot exceed $C^r$. These limited capacities result in a restriction on the order quantities of the warehouse and retailers. Additionally, at time point $t_4$,  the orders of the warehouse and retailers are not delivered, so the on-hand inventory and pipeline inventory of the warehouse and retailers do not change, i.e., $I^w_{t_4}=I^w_{t_3}$, $I^i_{t_4}=I^i_{t_3},i=1,\ldots,K.$

\par \textbf{Delivery and Update:} At the end of the period, denoted by the virtual time point $t+1$, the orders placed during this period enter the pipeline inventory, and the pipeline inventory advances by one period. The on-hand inventory and pipeline inventory of the warehouse and retailers are then updated as follows:
\begin{align*}
I^w_{t+1}&=I^w_{t_4}-\sum_{i=1}^K q^i_t,\\
I^i_{t+1}&=I^i_{t_4}, i=1,\ldots,K,\\
Q^w_{t+1}&=(q^w_{t-l_w+1},\ldots,q^w_{t}),\\
Q^i_{t+1}&=(q^i_{t-l_r+1},\ldots,q^i_{t}), i=1,\ldots,K.
\end{align*}

\par Inventory held at the warehouse or retailers incurs a holding cost with a cost rate of $h_w$ and $h_r$, respectively. The total holding cost at period $t$ can be calculated as follows:
\begin{equation}
\text{holding cost: }h_w I^w_{t+1}+h_r \sum_{i=1}^K I^i_{t+1}.
\label{holding_cost}
\end{equation}

\par
\begin{figure}[htb]
{
\centering
\includegraphics[width=0.8\textwidth]{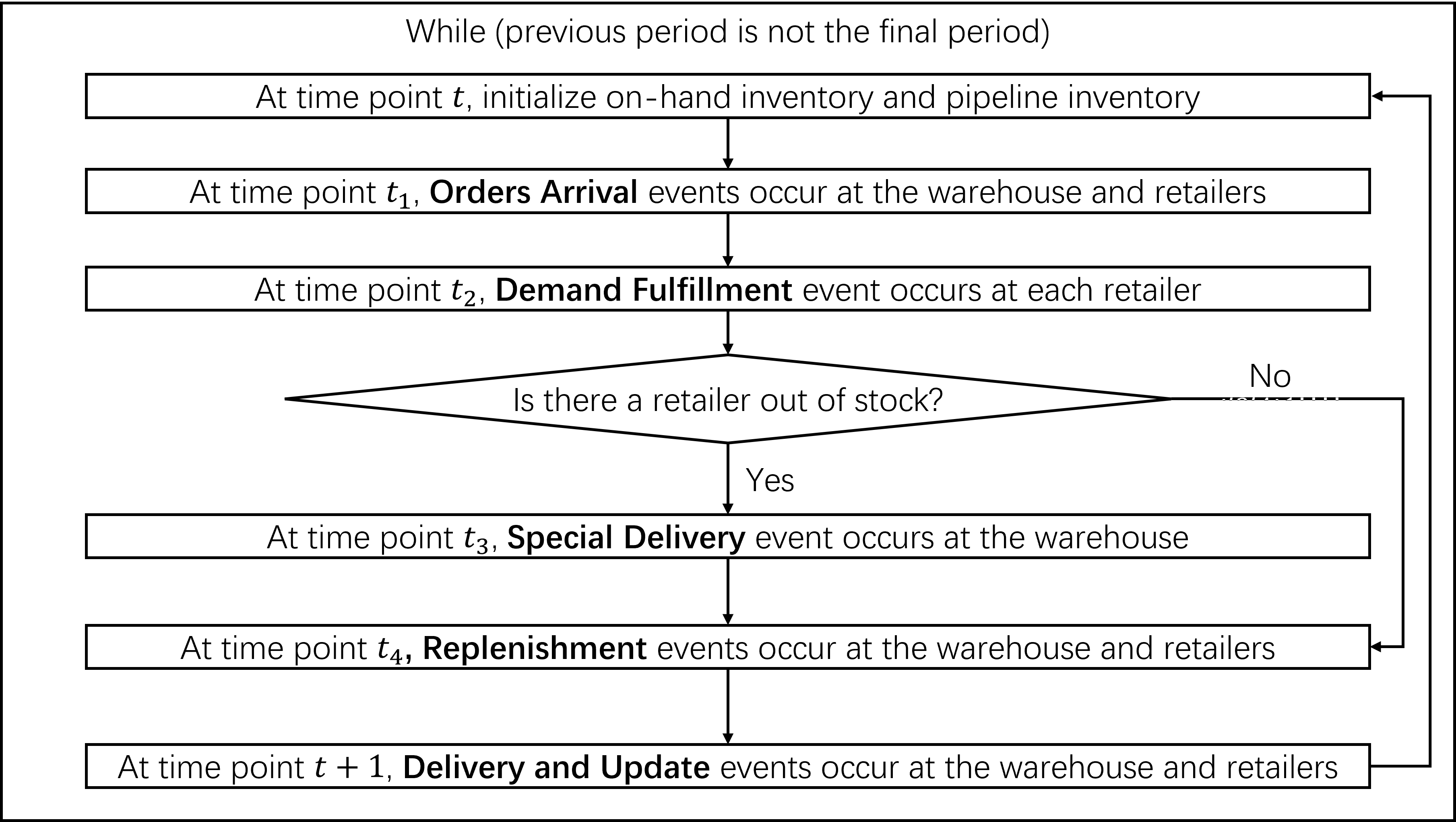}
\caption{The procedure of multi-echelon inventory management simulation for one period between time point $t$ and $t+1$.\label{procedure}}
}
\end{figure}

 To implement the simulation model, the sequence of these events is specified, and they occur in the following order at each period: $t<t_1<t_2<t_3<t_4<t+1$. The procedure of multi-echelon inventory management simulation for one period are illustrated in Figure \ref{procedure}.

\subsection{Markov Decision Process}

\par The multi-echelon system's dynamics can be formulated as a MDP. In this MDP, state $s_t=(I^w_t,Q^w_t,I^r_t,Q^r_t)$, where $(I^w_t,Q^w_t)$ are warehouse on-hand and pipeline inventory at period $t$, and $(I^r_t,Q^r_t)$ are retailers on-hand and pipeline inventory at period $t$. Action $a_t=(q^w_t,q^r_t)$, where $(q^w_t,q^r_t)$ are warehouse and retailers order quantities at time $t$. The reward in this MDP is represented by the cost, denoted as $c_t(s_t,a_t)$. The cost can be broken down into three parts, shortage cost, ordering cost, and holding cost, which are defined in \eqref{shortage_cost}, \eqref{ordering_cost} and \eqref{holding_cost}, respectively. The objective of this MDP is to minimize the expected cumulative costs by controlling actions from period $t$ to infinity:

\begin{equation}
\begin{aligned}
\min_{a_{t+j},j=0,1,\ldots} \sum_{j=0}^\infty \gamma^{j} E[c_{t+j}(s_{t+j},a_{t+j})],
\end{aligned}
\label{future_cost}
\end{equation}
where $\gamma$ is a discount rate. With a transition probability $P(s_{t+1} \mid s_t,a_t)$, which describes the probability of the system to transit from state $s_{t}$ to state $s_{t+1}$ when picking action $a_t$, the Objective \eqref{future_cost} can be achieved using linear programming or dynamic programming \cite{1998Reinforcement}.

\par Q-learning is an another approach for solving the problem. In Q-learning, the action-value function $Q(s_t,a_t)$ represents the future expected cost of taking action $a_t$ at state $s_t$ and then following the optimal control from state $s_{t+1}$. For any state $s_t \in S$, the action-value function can be written as:

$$
Q(s_t,a_t)=E[c_t(s_t,a_t)]+\min_{a_{t+j},j=1,2,\ldots} \sum_{j=1}^\infty \gamma^{j} E[c_{t+j}(s_{t+j},a_{t+j} )].
$$

\par The Objective \eqref{future_cost} can be achieved equivalently by minimizing $Q(s_t,a_t)$. Thus the optimal action can be calculated by:
$$
a^{\star}_t=\arg\min_{a_t}{Q(s_t,a_t)}.
$$

\par The Q-learning approach begins by assigning an initial Q-value, typically set to 0, to all states and actions. It then iteratively update the Q-values using the following formula:
$$
Q_{t+1}(s_t,a_t)=(1-\alpha)Q_{t}(s_t,a_t)+\alpha E[c_t(s_t,a_t)+\gamma \min_{a}Q_t(s_{t+1},a)],
$$
where $\alpha$ is the learning rate. The agent chooses actions using the $\epsilon$-greedy method, which implies that the agent chooses an action randomly with a probability $\epsilon$, and selects the action with the smallest Q-value with a probability of $1-\epsilon$.

\par
\begin{figure}[htb]
{
\centering
\vspace{0.5cm}
\includegraphics[width=0.8\textwidth]{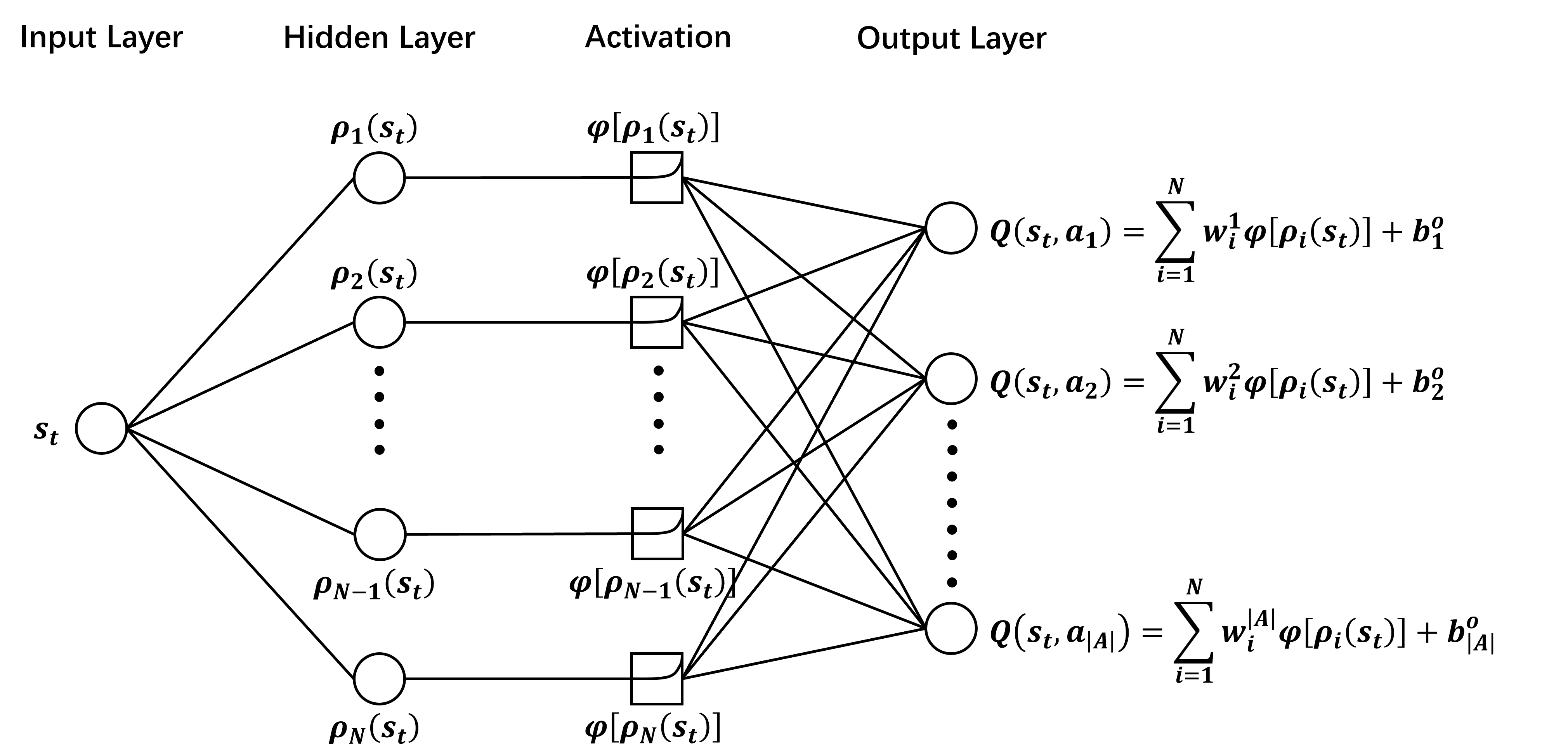}
\caption{Structure of a deep Q-network. In RBF based deep Q-network, $\rho_i(s_t)=\left\| s_t - s_i \right\|$ is Euclidean distance and activation function $\varphi[\rho_i(s_t)]=k(\left \|  s_t - s_i\right \|)$ is kernel function. While in deep Q-network constructed by other neural networks, $\rho_i(s_t)=\theta^T_i s_t +b^h_i$ is a linear transformation of its inputs $s_t$ and activation function is typically the sigmoid function or the Rectified Linear Unit (ReLU) function.\label{DQN}}
}
\end{figure}

 In a typical Q-learning process, the values of $Q(s_t,a_t)$ are stored in a lookup table, called Q-table. Solving MDP with a large state-action space by updating the Q-table values is impossible, which is known as the curse of dimensionality. To address this, \shortciteN{2015Human} developed a deep Q-network (DQN) algorithm that uses a neural network as an approximation function $\hat{Q}(s_t,a_t;W_t)$ of action-value function $Q(s_t,a_t)$. Figure \ref{DQN} shows the structure of a typical deep Q-network, which has $|A|$ outputs in the output layer representing the approximated values of $Q(s_t,a_t)$ for every possible action $a_t \in A$. Based on this structure, we propose a specialized and easy-to-design deep Q-Network for solving the multi-echelon inventory management MDP.

\section{Deep Q-Network Based On Radial Basis Functions}

A radial basis function (RBF) network is used to construct the deep Q-network. The RBF based deep Q-network is a special three-layers network with $N$ hidden neurons corresponding to states $\{ s_1, s_2, \ldots, s_N\}$. As illustrated in Figure \ref{DQN}, the output of each hidden neuron is given by $\rho_i(s_t)=\left\| s_t - s_i \right\|$, which is the Euclidean distance between the current state $s_t$ and the hidden neuron $s_i$. This is a key difference between the RBF based deep Q-network and Q-networks with other neural network architectures, such as fully connected neural networks (FCNs) or convolutional neural networks (CNNs), where the output of each hidden neuron is typically a linear transformation of its inputs $s_t$, given by $\rho_i(s_t)=\theta^T_i s_t +b^h_i$. Because the hidden layer neurons in the RBF based deep Q-network correspond to states, we can select lattice points from the minimum state to the maximum state to cover the entire state space. This makes the RBF based deep Q-network easy to design and implement in two ways. First, we don't need to determine the structure of hidden layers, such as the number of hidden layers and the number of neurons per hidden layer. Second, the real meaning of hidden layer neurons provides guidance for construction, in contrast to other neural network architectures which are primarily constructed based on intuition and experience. Overall, by using an RBF network to construct the deep Q-network, we aim to simplify the design and implementation process while leveraging the unique properties of RBF networks to effectively approximate the action-value function in the multi-echelon inventory management MDP.

\par Another key difference between RBF based deep Q-network and deep Q-networks with other neural network architectures is the activation function. In RBF based deep Q-network, the activation function $\varphi[\rho_i(s_t)]=k(\left \|  s_t - s_i\right \|)$ is a kernel function that converts the Euclidean distance $\left \| s_t - s_i\right \|$ to a high dimensional space distance, while common activation functions used in other deep Q-networks are the sigmoid function or the Rectified Linear Unit (ReLU) function. Thus, each output of the RBF based deep Q-network, as shown in Figure \ref{DQN}, is a linear combination of $N$ kernel functions, where each kernel function measures the high dimensional space distance between the current state $s_t$ and the hidden neuron $s_i$. The most widely used kernel function is the radial basis kernel function, which is why it is called an "radial basis kernel" network. The radial basis kernel function is defined as:
$$k(s_t,s_i)=\exp{(-\frac{{\left \|  s_t - s_i\right \|}^2}{2\eta^2})}.$$
The literature also suggests other kernel functions, for example, the Matérn$(\nu)$ kernel function. With gamma function $\Gamma(\cdot)$ and the modified Bessel function $K_{\nu}(\cdot)$, the Matérn$(\nu)$ kernel function is:

$$
k_{\text {Matérn }(\nu)}\left(s_t,s_i \right):=\frac{1}{2^{\nu-1} \Gamma(\nu)}\left(\sqrt{2 \nu}\left\|\eta^{\top}\left(s_t,s_i\right)\right\|\right)^{\nu} K_{\nu}\left(\sqrt{2 v}\left\|\eta^{\mathrm{\top}}\left(s_t,s_i\right)\right\|\right).
$$
 The Matérn kernel has a simplified form if $\nu$ is a half-integer: $\nu=p+\frac{1}{2}$ for some non-negative
integer $p$, and the Matérn kernel becomes more differentiable as $p$ increases. For instance, Matérn$(\frac{5}{2})$ kernel function is second-order differentiable:
\begin{equation}
k_{\text {Matérn }(\frac{5}{2})}\left(s_t,s_i \right):= \left( 1+\frac{\sqrt{5}\left\| s_t - s_i \right\|}{\eta}+\frac{5\left\| s_t - s_i \right\|^{2}}{3 \eta^{2}} \right ) \exp \left(-\frac{\sqrt{5}\left\|s_t - s_i\right\|}{\eta}\right).
\label{matern_eq}
\end{equation}
$\eta$ is a hyperparameter that determines the RBF based deep Q-network's smoothness. A smaller $\eta$ results in a less smooth RBF based deep Q-network.

\par After constructing the RBF based deep Q-network, we train it by minimizing loss function $L(W)$, which is derived from Q-learning:
\begin{equation}
		L(W)=\left ( E[c_t(s_t,a_t)]+ \gamma \min_a \hat{Q}(s_{t+1},a;W)  - \hat{Q}(s_t,a_t;W) \right )^{2}.
\end{equation}

\par The loss function measures the difference between the current action value $\hat{Q}(s_t,a_t;W)$ and the predicted action value $E[c_t(s_t,a_t)]+ \gamma \min_a \hat{Q}(s_{t+1},a;W)$. The loss function is minimized via the gradient descent method. Considering a learning rate $\alpha$, the weight vector $W$ is updated by:
\begin{equation}
		W_{t+1}= W_{t} + \alpha \left ( E[c_t(s_t,a_t)]+ \gamma \min_a \hat{Q}(s_{t+1},a;W_t)  - \hat{Q}(s_t,a_t;W_t) \right ) \nabla \hat{Q}(s_t,a_t;W_t).
\end{equation}
where $\nabla \hat{Q}(s_t,a_t;W_t)$ is the gradient of $\hat{Q}(s_t,a_t;W_t)$. As each output of RBF based deep Q-network is a combination of $N$ kernel functions, it is easy to derive that the gradient is $\left ( k(s_t,s_1), k(s_t,s_2), \ldots, k(s_t,s_N)\right )$, a vector of $N$ kernel function values.

\par Once the RBF based deep Q-network $\hat{Q}(s_t,a_t;W)$ is trained, the optimal order quantities at state $s_t$ are selected by minimizing $\hat{Q}(s_t,a_t;W)$:
\begin{equation}
		a^{\star}_t=\arg\min_{a_t}{\hat{Q}(s_t,a_t;W)}
\end{equation}

\section{Simulation Study}

This section evaluates the performance of the proposed RBF based deep Q-network in three numerical experiments. The first experiment is a simple serial system with one warehouse and one retailer, while the other two experiments are complex systems involving multiple retailers. The difference between the two complex systems is that the third system's demands are more unstable than the second system, with a larger demand standard deviation and longer lead times. Previous studies have explored these systems using different DRL approaches: \shortciteN{1997Aneuro} developed a neuro-dynamic programming approach for all three systems, and \shortciteN{2022Can} applied the A3C algorithm to study the two complex systems. We adopt the same settings as these studies to compare with their DRL approaches.

\par Similar to \shortciteN{1997Aneuro} and \shortciteN{2022Can}, we apply a baseline method and compare our approach's improvements against it. The baseline method is the base-stock policy, which means that for an installation (warehouse or retailer) with a base-stock level $s$, if the inventory position is less than $s$, the installation places orders to increase the inventory position to $s$ as close as possible. It should be noted that in a serial system, the base-stock policy is optimal \cite{1960Optimal}, while in a complex multi-echelon system, the optimal policy is unknown.

\par We select lattice states from the minimum state to the maximum state in hidden layers. The higher the state dimension, the more hidden layer neurons are used. Therefore, we reduce the state's dimension to reduce the hidden layer neurons. We set the state $s_t=({Z}^w_t,{Z}^r_t)$, where ${Z}^w_t=I^w_t+\sum_{j=1}^{l_w}q^w_{t-j}$ is the warehouse's inventory position, and ${Z}^r_t=\sum_{i=1}^K (I^i_t+\sum_{j=1}^{l_r}q^i_{t-j})$ is the total inventory position of all retailers. Thus, we reduce the states to two dimensions. We also reduce actions to two dimensions in a similar way. Actions are $a_t=({q}^w_t,{q}^r_t)$, where ${q}^w_t$ is the warehouse's order quantity, and ${q}^r_t$ is every retailer's order quantity, which are the same for all retailers.

\par The kernel function in our RBF based deep Q-network is the Matérn$(\frac{5}{2})$ kernel given by \eqref{matern_eq}. The hyperparameter $\eta$ that determines the RBF based deep Q-network's smoothness is set to 1 for all three experiments, suggesting that the RBF based deep Q-network is very unsmooth.

\par The simulation programs for the three experiments are implemented in C++. Details of the procedures are discussed in Section 2.1. The programs for the RBF-based deep Q-network algorithm are implemented in Python, and the ctypes library is used to call the C++ simulation programs. All experiments run on a 64-bit Linux machine with a 20$\times$2.50GHz CPU and 12$\times$16GB RAM.

\subsection{Experiment With One Warehouse One Retailer Serial System}

In the first experiment, the system consists of only one warehouse and one retailer. Furthermore, there is no lead time for the warehouse, and the retailer has only one period lead time. A detailed list of parameters is presented in Table \ref{tab_first}.

\begin{table}[htb]
\centering
\caption{Settings of a serial system with one warehouse and one retailer.\label{tab_first}}
\begin{tabular}{|c|c|c|c|c|c|c|c|c|c|c|c|c|c|}
\hline
& $\mu$ & $\sigma$ & $l_w$ & $l_r$ & $K$ & $h_w$ & $h_r$ & $c_w$ & $p$ & $P_w$ & $C^m$ & $C^w$ & $C^r$  \\ \hline
Setting 1 & 5 & 8 & 0 & 1 & 1 & 1 & 2 & 10 & 50 & 1 & 10 & 50 & 50\\
\hline
\end{tabular}
\end{table}

\par We selected lattice states as $\{(Z^w_i,Z^r_i): Z^w_i=0,5,\ldots,50, Z^r_i=0,5,\ldots,50\}$. Since they are also hidden layer neurons, the hidden layers have $N=121$ neurons. Regarding actions, we set the warehouse order quantity $q^w_t \in [0,10]$ and retailer order quantity $q^r_t \in [0,10]$. There are $121$ possible actions in total, implying that the output layer's dimension is $121$.

\par
\begin{figure}[htb]
{
\centering
\includegraphics[width=0.6\textwidth]{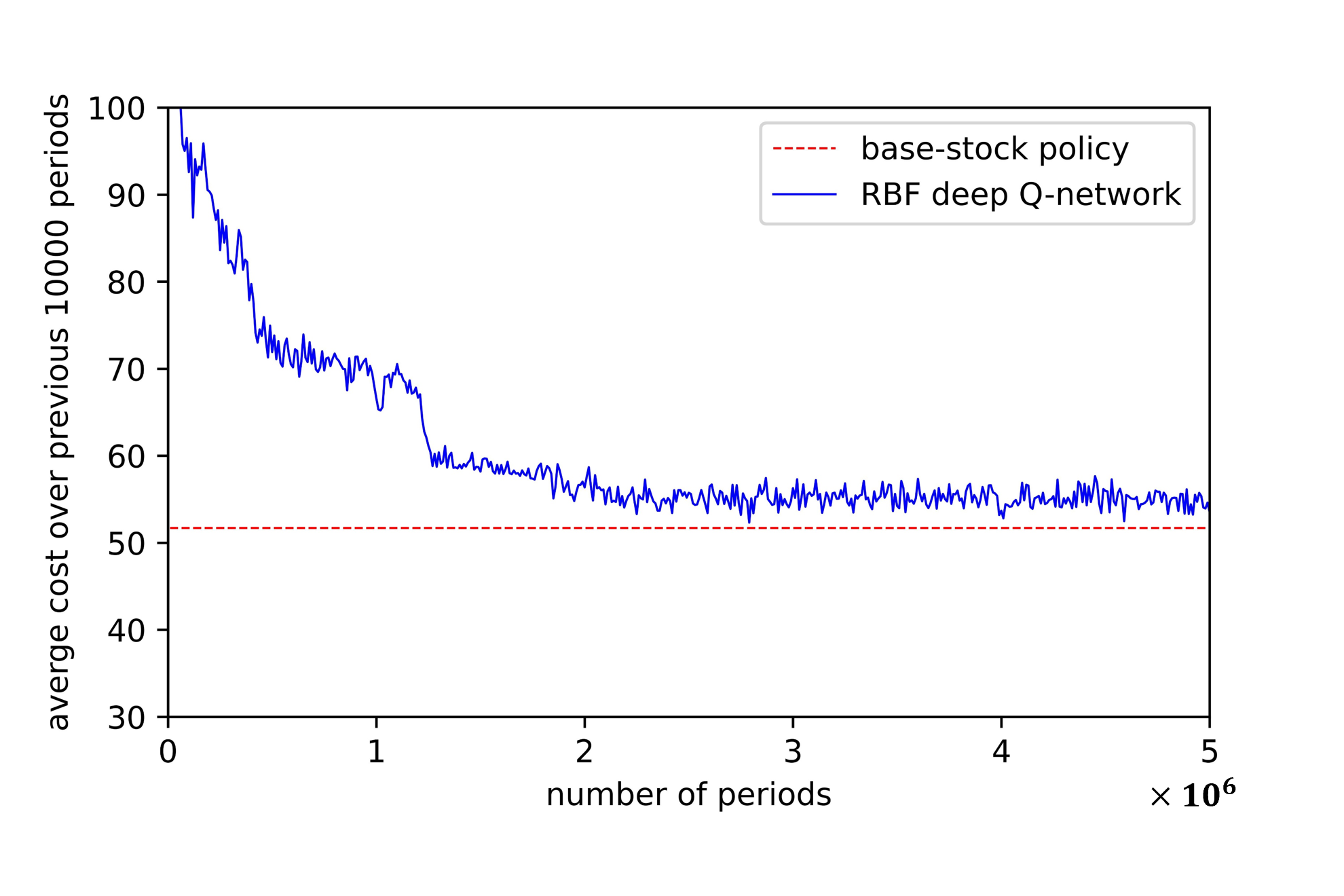}
\caption{Average cost evolution during training.\label{converge_1rdc}}
}
\end{figure}

 Figure \ref{converge_1rdc} displays the average cost evolution during the training process, which takes a total of 1,837 seconds. As shown, the average cost of the RBF based deep Q-network (blue solid line) decreases in the first 2,000,000 periods and stabilizes near the cost of the base-stock policy (red dashed line). There is a slight gap between the RBF based deep Q-network and the base-stock policy in the serial system, where the base-stock policy is optimal.

\par We also compare the on-hand inventory of base-stock policy and RBF based deep Q-network. Table \ref{tab_second} report the average warehouse and retailer on-hand inventory of base-stock policy $(\Bar{I}^w_{\text{BS}},\Bar{I}^r_{\text{BS}})$ and the average warehouse and retailer on-hand inventory of RBF based deep Q-network $(\Bar{I}^w_{\text{RBF}},\Bar{I}^r_{\text{RBF}})$, respectively. As shown in the table, their on-hand inventory is similar, especially for the retailer on-hand inventory. Moreover, the average relative difference between their on-hand inventory is also reported in Table \ref{tab_second}, and the difference is small. This indicates that the RBF based deep Q-network approach not only reduces costs to a near-optimal level but also orders and controls on-hand inventory like the base-stock policy. Thus, the RBF based deep Q-network learns a near-optimal solution in the serial system.

\begin{table}[htb]
\centering
\caption{Average on-hand inventory of base-stock policy and RBF based deep Q-network, and average relative difference between their on-hand inventory.\label{tab_second}}
\begin{tabular}{|c|c|c|c|c|c|}
\hline
$\Bar{I}^w_{\text{BS}}$ & $\Bar{I}^w_{\text{RBF}}$ & $\Bar{I}^r_{\text{BS}}$ & $\Bar{I}^r_{\text{RBF}}$ & Average of $\frac{| I^w_{\text{BS}}-I^w_{\text{RBF}} |}{I^w_{\text{BS}}}$ & Average of $\frac{| I^r_{\text{BS}}-I^r_{\text{RBF}} |}{I^r_{\text{BS}}}$ \\ \hline
4.71 & 6.05 & 13.28 & 12.28 & 34.88~\% & 9.66~\% \\
\hline
\end{tabular}
\end{table}

\par We calculate relative gap between RBF based deep Q-network and base-stock policy, and compare the result with \shortciteN{1997Aneuro}. Their approach has a 1.74\% gap to base-stock policy, while our gap is 2.87\%. It should be noted that \shortciteN{1997Aneuro} manually developed three features of the system as state, while our approach can achieve a similar near-optimal solution without manual feature engineering.

\subsection{Experiment With One Warehouse Multiple Retailers Multi-echelon System}

Next, we evaluate our approach in two systems, both with one warehouse and $K$ identical retailers. Table \ref{tab_third} lists the parameter settings. In the system with Setting 2, the demands are more stable, with a smaller demand standard deviation and shorter lead times. In the system with Setting 3, the demands are very unstable, with a demand mean of zero and a very large demand standard deviation.

\begin{table}[htb]
\centering
\caption{Settings of one warehouse and multiple retailers system.\label{tab_third}}
\begin{tabular}{|c|c|c|c|c|c|c|c|c|c|c|c|c|c|}
\hline
& $\mu$ & $\sigma$ & $l_w$ & $l_r$ & $K$ & $h_w$ & $h_r$ & $c_w$ & $p$ & $P_w$ & $C^m$ & $C^w$ & $C^r$  \\ \hline
Setting 2 & 5 & 14 & 2 & 2 & \multirow{2}{*}{10} & \multirow{2}{*}{3} & \multirow{2}{*}{3} & \multirow{2}{*}{0} & \multirow{2}{*}{60} & \multirow{2}{*}{0.8} & \multirow{2}{*}{100} & \multirow{2}{*}{1,000} & \multirow{2}{*}{100}\\
\cline{1-5} Setting 3 &  0 & 20 & 5 & 3 & & & & & & & & & \\
\hline
\end{tabular}
\end{table}

\par The selected lattice states (hidden neurons) for Setting 2 are $\{(Z^w_i,Z^r_i): Z^w_i=200,220,\ldots,400, Z^r_i=100,120,\ldots,400\}$, and for Setting 3 are $\{(Z^w_i,Z^r_i): Z^w_i=300,320,\ldots,600, Z^r_i=100,120,\ldots,300\}$. Each setting has $N=176$ neurons in the hidden layer. The action ranges are set to $q^w_t \in [50,100]$ and $q^r_t \in [0,15]$ in Setting 2, while $q^w_t \in [40,100]$ and $q^r_t \in [0,15]$ in Setting 3. The output layer's dimensions are $816$ and $976$ in Setting 2 and Setting 3, respectively.

\par
\begin{figure}[htb]
	\centering  
	\subfigure[Average cost of Setting 2]{
		\includegraphics[width=0.48\linewidth]{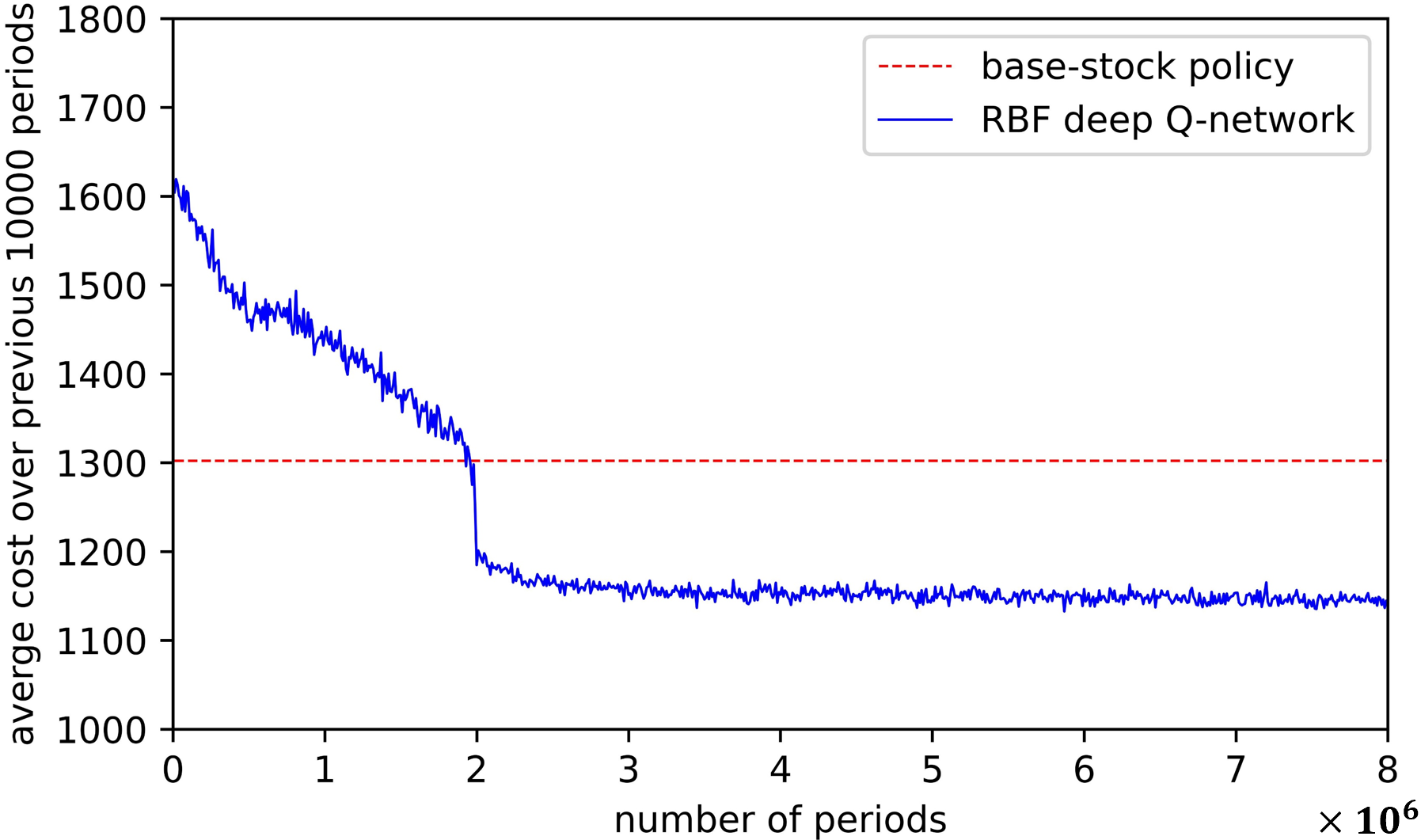}}
	\subfigure[Average cost of Setting 3]{
		\includegraphics[width=0.48\linewidth]{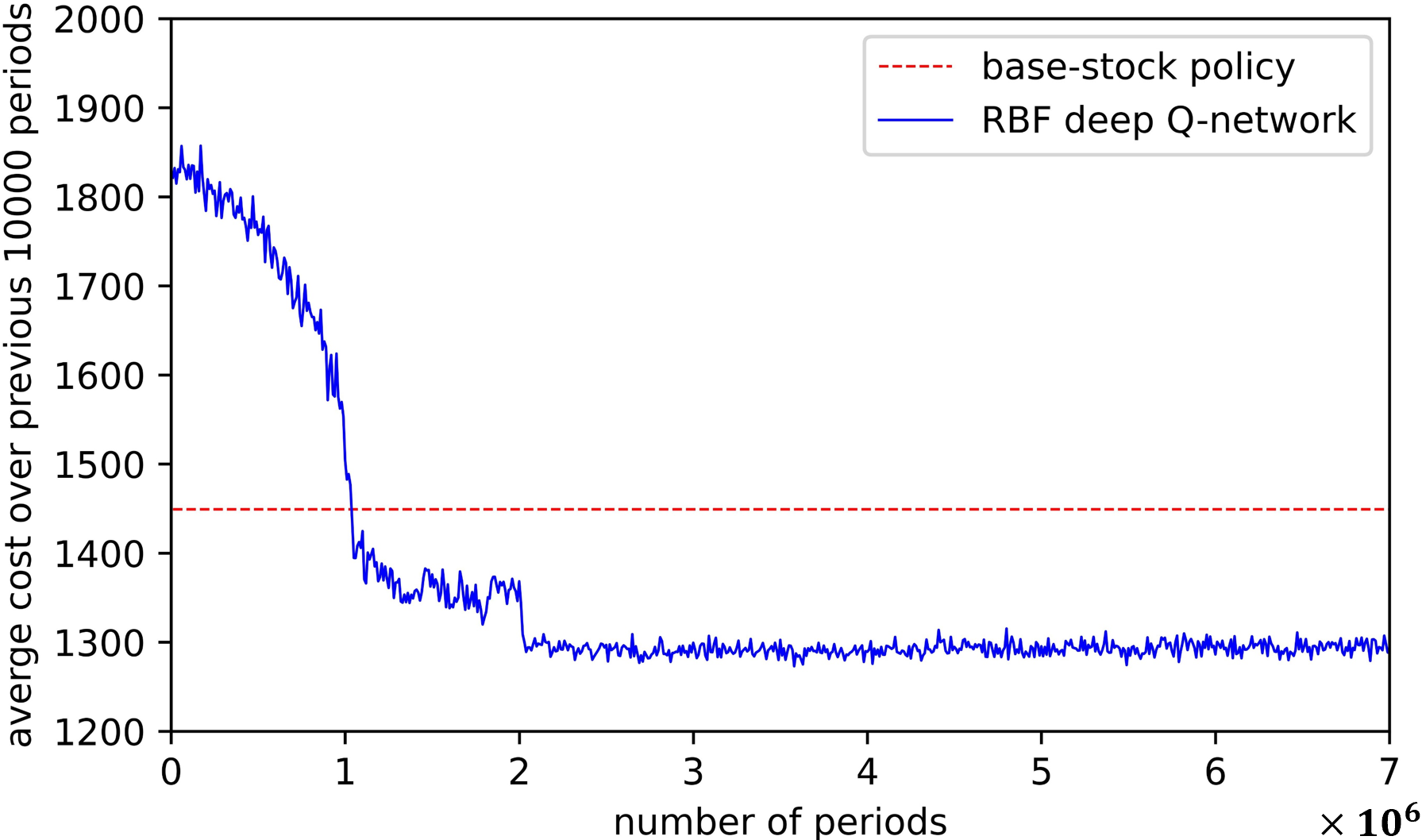}}
	\caption{Average cost evolution during training.}
 \label{converge_case12}
\end{figure}

 Figure \ref{converge_case12} illustrates the average cost evolution of Setting 2 and Setting 3 during training, which takes a total of 3,832 seconds and 4,656 seconds, respectively. In both figures, the average costs of the RBF based deep Q-network (blue solid lines) initially reduce, then stabilize, and finally become lower than the base-stock policy costs (red dashed lines). This implies that the base-stock policy is no longer optimal for complex multi-echelon systems with multiple retailers.

\par We also compare the on-hand inventory of base-stock policy and RBF based deep Q-network. Table \ref{tab_forth} reports the average warehouse and retailer on-hand inventory of base-stock policy $(\Bar{I}^w_{\text{BS}},\Bar{I}^r_{\text{BS}})$ and the average warehouse and retailer on-hand inventory of RBF based deep Q-network $(\Bar{I}^w_{\text{RBF}},\Bar{I}^r_{\text{RBF}})$ in Setting 2 and Setting 3. In both settings, the average warehouse on-hand inventory of RBF based deep Q-network is lower than the base-stock policy. This implies warehouses controlled by RBF based deep Q-network order less and can achieve lower warehouse holding costs. Regarding the average retailer on-hand inventory, in Setting 2, the RBF based deep Q-network and base-stock policy are the same, while in Setting 3, the base-stock policy is lower.

\begin{table}[htb]
\centering
\caption{Average on-hand inventory of base-stock policy and RBF based deep Q-network in two settings.\label{tab_forth}}
\begin{tabular}{|c|c|c|c|c|c|c|}
\hline
& $\Bar{I}^w_{\text{BS}}$ & $\Bar{I}^w_{\text{RBF}}$ & $\Bar{I}^r_{\text{BS}}$ & $\Bar{I}^r_{\text{RBF}}$ \\ \hline
Setting 2 & 157 & 109 & 106 & 106 \\
\hline
Setting 3 & 154 & 123 & 110 & 134 \\
\hline
\end{tabular}
\end{table}

\par It is worth noting that we find in both Setting 2 and Setting 3, the larger the warehouse on-hand inventory, the more retailers will order. This reduces warehouse holding costs because once the warehouse on-hand inventory is large, the retailers will order more to reduce warehouse on-hand inventory. This also implies that a retailer controlled by RBF based deep Q-network considers both the warehouse and retailer inventory when making decisions, while a retailer following the base-stock policy only considers its own inventory to make decisions. Hence, the RBF based deep Q-network achieves lower costs by learning more information.

\par Furthermore, we calculate the relative improvement of RBF based deep Q-network compared to the base-stock policy and compare our relative improvement with current DRL approaches. Table \ref{tab_fifth} shows the relative improvement of different DRL approaches. The RBF based deep Q-network is slightly better than both neuro-dynamic programming and A3C in Setting 2 and is as good as A3C in Setting 3. It's worth noting that \shortciteN{1997Aneuro} manually developed 23 features for the neuro-dynamic programming approach, while both the RBF based deep Q-network and A3C do not require manual feature engineering. Additionally, the process of designing the neural network in A3C is complex. As pointed out by \shortciteN{2022Can}, tuning to select the number of hidden layers and neurons per layer remains computationally burdensome. The tuning and training time for A3C can be days or even weeks. In contrast, the RBF based deep Q-network does not require special design for the neural network structure. Thus, the RBF based deep Q-network is easier to implement. The training process for Setting 2 and Setting 3 takes only 3832 seconds and 4656 seconds, respectively, which is significantly less than the A3C algorithm.

\begin{table}[htb]
\centering
\caption{Relative improvement of different DRL approaches.\label{tab_fifth}}
\begin{tabular}{|c|c|c|c|}
\hline
& RBF Based Deep Q-network & \makecell{Neuro-dynamic Programming \\ \shortcite{1997Aneuro}} & \makecell{A3C \\ \shortcite{2022Can}}\\ \hline
Setting 2 & 12~\% & 10~\% & 9~\%\\
\hline
Setting 3 & 12~\% & 10~\% & 12~\%\\
\hline
\end{tabular}
\end{table}

\section{Conclusion}

This paper proposes a deep Q-network approach based on RBF to solve dynamic inventory management for general multi-echelon systems. The RBF based deep Q-network has a simple structure and can be easily constructed. Simulation studies show that our method performs better than the base-stock policy in multi-echelon systems with multiple retailers. Meanwhile, we also compared the RBF based deep Q-network with current DRL approaches and find that the RBF based deep Q-network has appealing performance compared to existing DRL approaches and is easier to design. These demonstrate the potential use of our RBF based deep Q-network for solving practical inventory management problems.


\section*{Acknowledgement}
This work was partially supported by the National Natural Science Foundation of China [No. 72031006, 72071146] and Alibaba Innovative Research (AIR) Project.

\footnotesize

\bibliographystyle{wsc}
\bibliography{reference}

\section*{AUTHOR BIOGRAPHIES}

\noindent {\bf LIQIANG CHENG} is a master student in the Antai College of Economics and Management at Shanghai Jiao Tong University. His email address is \email{chengzi1998@sjtu.edu.cn}.\\

\noindent {\bf JUN LUO} is a professor in the Antai College of Economics and Management at Shanghai Jiao Tong University. His primary research interest are simulation optimization and statistics. His email address is \email{jluo\_ms@sjtu.edu.cn}.\\

\noindent {\bf WEIWEI FAN} is an associate professor in the Advanced Institute of Business and School of Economics and Management at Tongji University. Her primary research interest are simulation optimization and robust optimization. Her email address is \email{wfan@tongji.edu.cn}.\\

\noindent {\bf YIDONG ZHANG} is a director of the B2C Supply Chain Optimization team and the Supply Planning Optimization team at Alibaba Group. His email address is \email{tanfu.zyd@alibaba-inc.com}.\\

\noindent {\bf YUAN LI} is an algorithm engineer in the Dchain Department at Alibaba Group. His email address is \email{yuan.lya@alibaba-inc.com}.
\end{document}